\RequirePackage{fix-cm}
\documentclass[10pt]{article}
\usepackage[a4paper, total={6in, 8in}]{geometry}

\usepackage{graphicx}
\usepackage{xcolor}
\usepackage{
        ,latexsym
        ,algorithm
	,algorithmicx,
	,algpseudocode
        ,amsmath
        ,amssymb
        ,amsfonts
        }

\usepackage{tikz}
\usetikzlibrary{positioning}
\usetikzlibrary{decorations.text}
\usetikzlibrary{arrows}
\usetikzlibrary{shapes}

\usepackage{caption}
\usepackage{pgfplotstable} 
\usepackage[utf8]{inputenc}

\usepackage{booktabs}
\usepackage{graphicx}

\usepackage{pgfplots}

\usepackage{amsmath}
\usepackage{amsthm}

\usepackage{xcolor}

\newcommand{\gstar}{g_{*}}
\newcommand{\gconj}{g_{\wedge}}
\newcommand{\gdisj}{g_{\vee}}

\newtheorem{example}{Example}

\title{Stacked Structure Learning for \\ Lifted Relational Neural Networks}

\author{Gustav \v{S}ourek\footnote{Faculty of Electrical Engineering, CTU in Prague, Czech Republic, Email: souregus@fel.cvut.cz} \and Martin Svato\v{s}\footnote{Faculty of Electrical Engineering, CTU in Prague, Czech Republic, Email: svatoma1@fel.cvut.cz} \and Filip \v{Z}elezn\'{y}\footnote{Faculty of Electrical Engineering, CTU in Prague, Czech Republic, Email: zelezny@fel.cvut.cz} \and  Steven~Schockaert\footnote{School of Computer Science, Cardiff University, UK, Email: SchockaertS1@cardiff.ac.uk} \and Ond\v{r}ej~Ku\v{z}elka\footnote{School of Computer Science, Cardiff University, UK, Email: KuzelkaO@cardiff.ac.uk}}

\begin{document}

\maketitle

\begin{abstract}
Lifted Relational Neural Networks (LRNNs) describe relational domains using weighted first-order rules which act as templates for constructing feed-forward neural networks. While previous work has shown that using LRNNs can lead to state-of-the-art results in various ILP tasks, these results depended on hand-crafted rules. In this paper, we extend the framework of LRNNs with structure learning, thus enabling a fully automated learning process. Similarly to many ILP methods, our structure learning algorithm proceeds in an iterative fashion by top-down searching through the hypothesis space of all possible Horn clauses, considering the predicates that occur in the training examples as well as invented soft concepts entailed by the best weighted rules found so far. In the experiments, we demonstrate the ability to automatically induce useful hierarchical soft concepts leading to deep LRNNs with a competitive predictive power.


\end{abstract}

\section{Introduction}

Lifted Relational Neural Networks (LRNNs \cite{Sourek2015}) are weighted sets of first-order rules, which are used to construct feed-forward neural networks from relational structures. A central characteristic of LRNNs is that a different neural network is constructed for each learning example, but crucially, the weights of these different neural networks are shared. This allows LRNNs to use neural networks for learning in relational domains, despite the fact that training examples may vary considerably in size and structure. 

In previous work, LRNNs have been learned from hand-crafted rules. In such cases, only the weights of the first-order rules have to be learned from training data, which can be accomplished using a variant of back-propagation. The use of hand-crafted rules offers a natural way to incorporate domain knowledge in the learning process. In some applications, however, (sufficient) domain knowledge is lacking and both the rules and their weights have to be learned from data. To this end, in this paper we introduce a structure learning method for LRNNs.

Our proposed structure learning method proceeds in an iterative fashion. In each iteration, it may either learn a set of rules that intuitively correspond to a new layer of a neural network template or to learn a set of rules that intuitively correspond to creating new connections among existing layers, a strategy which we refer to as stacked structure learning. The rules that are added in a given iteration either define one of the target predicates, or they define a new predicate that may depend on predicates that were `invented' at earlier layers as well as on predicates from the considered domain. Since the actual meaning of these predicates depends on both the learned rules and their associated weights, structure learning is alternated with weight learning. Intuitively, this means that the definitions of predicates defined in earlier layers can be fine-tuned based on the rules which are added to later layers.

We present experimental result which show that the resulting LRNNs perform comparably to LRNNs that have been learned from hand-crafted rules. We believe that this makes LRNNs a particularly convenient framework for learning in relational domains, without any need for prior knowledge nor for any extensive hypertuning. Somewhat surprisingly, we find that LRNNs with learned rules are often more compact than those with hand-crafted rules.

The remainder of the paper is structured as follows. In the next section, we first provide the required background on LRNNs. In Section \ref{sec:struct}, we then present the proposed stucture learning method, after which we discuss our experimental results in Section \ref{sec:experiments}.

\section{Preliminaries}\label{sec:preliminaries}

In this section, we briefly recall the LRNN framework from \cite{Sourek2015}.

\paragraph{{\bf LRNN Structure.}} A lifted relational neural network (LRNN) $\mathcal{N}$ is a set of weighted definite clauses, i.e.\ a set of pairs $(R_i,w_i)$ where $R_i$ is a definite clause and $w_i \in \mathbb{R}$. For a LRNN $\mathcal{N}$, we write $\mathcal{N}^*$ to denote the corresponding set of definite clauses, i.e. $\mathcal{N}^* = \{ C \,|\, (C,w) \in \mathcal{N} \}$.  
The grounding $\overline{\mathcal{N}}$ of a LRNN $\mathcal{N}$ is defined as $\overline{\mathcal{N}} = \{(C\theta,w) \,|\, (C,w)\in \mathcal{N}, C\theta\in G(\mathcal{N}^*)\}$, where $G(\mathcal{N}^*)$ is the restriction of the grounding of $\mathcal{N}^*$ to those clauses that correspond to active rules, i.e.\ rules whose antecedent is satisfied in the least Herbrand model of $\mathcal{N}^*$.
The neural network corresponding to $\overline{\mathcal{N}}$ contains the following types of neurons:
\begin{itemize}
\item For each ground atom $h$ occurring in $\overline{\mathcal{N}}$, there is a neuron $A_{h}$, called an {\em atom neuron}.
\item For each ground fact $(h,w) \in \overline{\mathcal{N}}$, there is a neuron $F_{(h,w)}$, called a {\em fact neuron}.
\item For every ground rule $(c\theta \leftarrow b_1\theta \wedge \dots \wedge b_k\theta,w) \in \overline{\mathcal{N}}$, there is a neuron $R_{(c\theta \leftarrow b_1\theta \wedge \dots \wedge b_k\theta,w)}$, called a {\em rule neuron}.
\item For every (possibly non-ground) rule $(c \leftarrow b_1 \wedge \dots \wedge b_k,w) \in \mathcal{N}$ and every grounding $h = c\theta$ of $c$ that occurs in $\mathcal{H}$, there is a neuron $\textit{Agg}_{(c \leftarrow b_1 \wedge \dots \wedge b_k,w)}^{h}$, called an {\em aggregation neuron}.
\end{itemize}

\begin{figure}[t]
\resizebox{\textwidth}{5.0cm}{
\begin{tikzpicture}
  \node (img1) {\includegraphics[width=\textwidth]{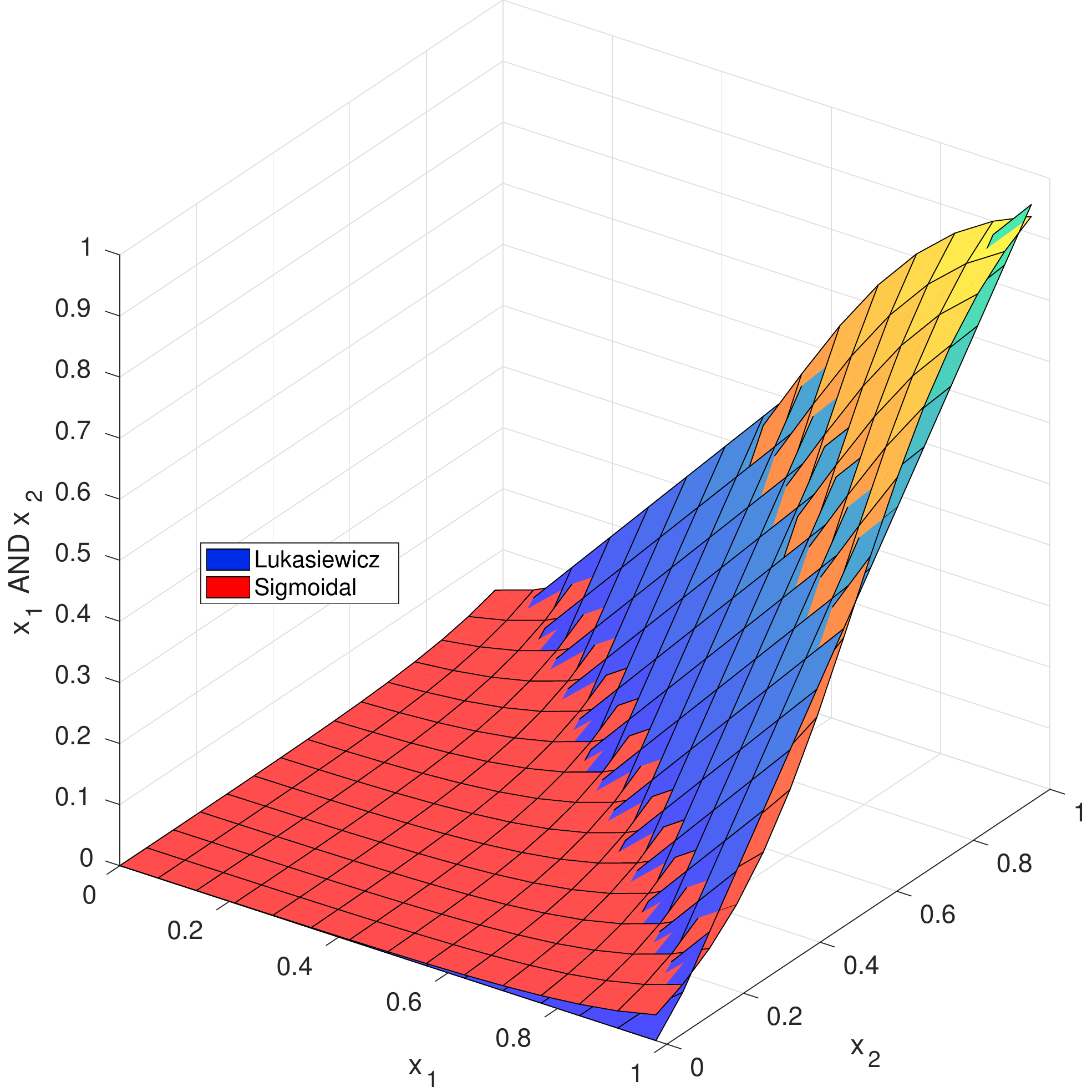}};
\end{tikzpicture}
\hspace{0.2cm}
\begin{tikzpicture}
  \node (img1) {\includegraphics[width=\textwidth]{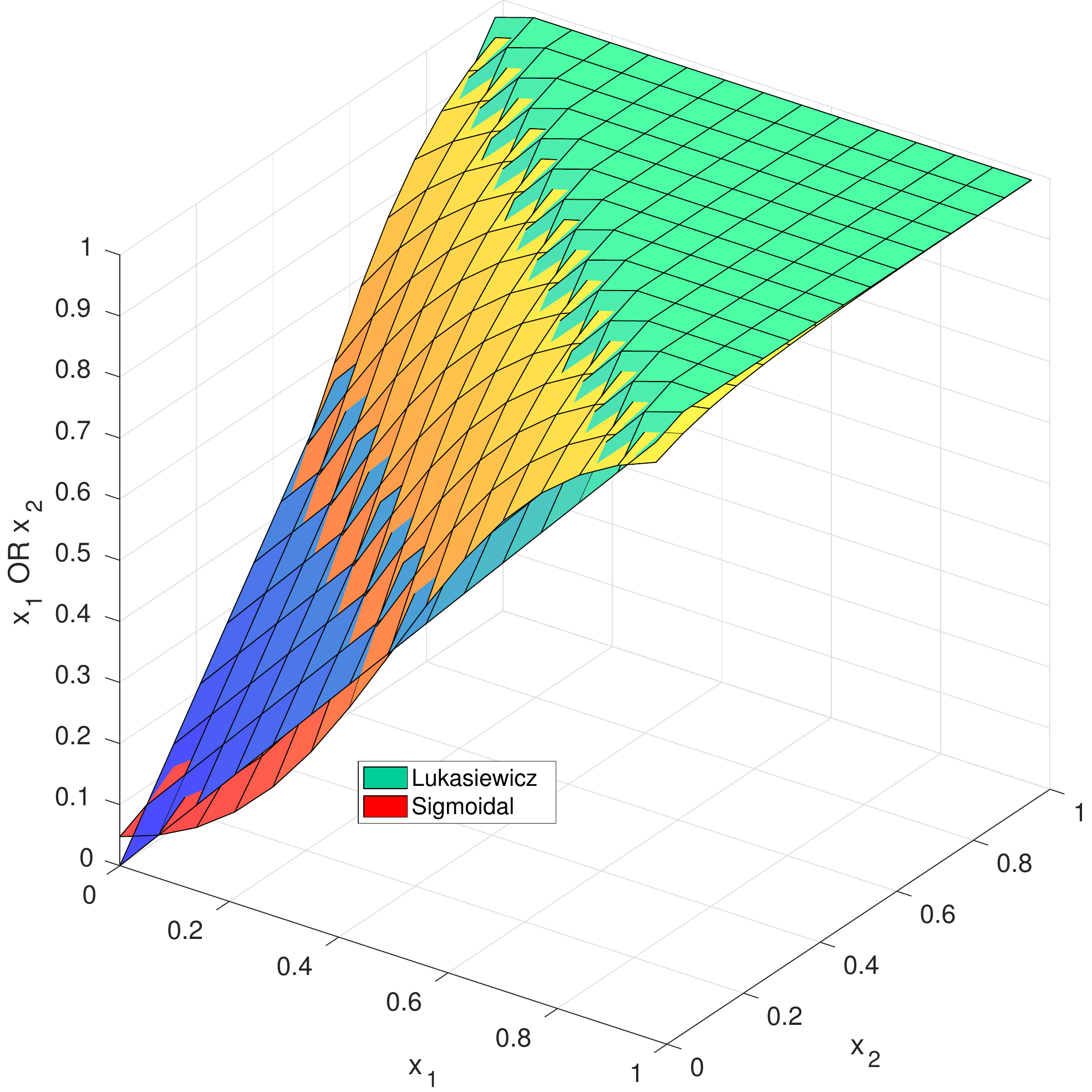}};
\end{tikzpicture}}
\caption{An approximation of {\L}ukasiewicz conjunction (left) and disjunction (right) by sigmoidal activation functions $\gconj$ and $\gdisj$ for the use in LRNNs.}
\label{fig:activations}
\end{figure}

\paragraph{{\bf Forward propagation.}} Intuitively, the neural network computes for each gro\-und atom $h$ a truth value, which is given by the output of the atom neuron $A_h$. To obtain these truth values, the network propagates values in a way which closely mimics the immediate consequence operator from logic progamming. In particular, when using the immediate consequence operator, there are two ways in which $h$ can become true: if $h$ corresponds to a fact, or if $h$ is the head of a rule whose body is already satisfied. Similarly, the inputs of the atom neuron $A_h$ consist of the fact neurons of the form $F_{(h,w)}$ and aggregation neurons of the form $\textit{Agg}_{(c \leftarrow b_1 \wedge \dots \wedge b_k,w)}^{h}$. The output of an atom neuron with inputs $i_1,...,i_m$ is given by $\gdisj(i_1,...,i_m)$, where $\gdisj$ is an activation function that maps the inputs to a real-valued output. In this paper we will use
$$\gdisj(b_1, \dots, b_k) =  \textit{sigm} \left( a \cdot \left( \sum_{i=1}^k b_i + b_0 \right) \right)$$
where $\textit{sigm}$ is the sigmoid function $\textit{sigm}(x) = 1/(1+e^{-x})$. We set the parameters $a=6$ and $b_0 = -0.5$, as $\gdisj$ then closely approximates the {\L}ukasiewicz fuzzy disjunction \cite{hajek1998metamathematics} (see right panel in Figure \ref{fig:activations}). This helps with the interpretability of LRNNs, as it means that we can intuitively think of the activation functions as logical connectives, and of LRNNs as (fuzzy) logic programs.

A fact neuron $F_{(h,w)}$ has no input and has the value $w$ as its output. The output of the aggregation neuron $\textit{Agg}_{(c \leftarrow b_1 \wedge \dots \wedge b_k,w)}^{h}$ intuitively expresses how strongly $h$ can be derived using the rule $c \leftarrow b_1 \wedge \dots \wedge b_k$. The inputs of the aggregation neuron $\textit{Agg}_{(c \leftarrow b_1 \wedge \dots \wedge b_k)}^{h}$ are all rule neurons $R_{(c\theta \leftarrow b_1\theta \wedge \dots \wedge b_k\theta,w)}$ for which $c\theta=h$. The output of this aggregation neuron is given by $w\cdot \gstar(i_1,...,i_m)$, where $i_1,...,i_m$ are its inputs, $\gstar$ is an activation function, and $w$ is the weight of the corresponding rule. We will use
$$\gstar(b_1, \dots, b_m) = \frac{1}{m}\sum_{i=1}^m b_i.$$

The rule neuron $R_{(c\theta \leftarrow b_1\theta \wedge \dots \wedge b_k\theta,w)}$ intuitively needs to fire if the atoms $b_1\theta,...,b_k\theta$ are all true. Accordingly, its inputs $i_1,...,i_k$ are given by the atom neurons $A_{b_1\theta},...,A_{b_k\theta}$, and its output is $\gconj(i_1,...,i_k,w)$, with $\gconj$ a third type of activation function.
In this paper we will use the activation function
$$\gconj(b_1, \dots, b_k) = \textit{sigm} \left( a \cdot \left( \sum_{i=1}^k b_i -k+1 + b_0 \right) \right)$$
where we set $a=6$ and $b_o = -0.5$, which approximates {\L}ukasiewicz fuzzy conjunction \cite{hajek1998metamathematics} (see left panel in Figure \ref{fig:activations}).

\paragraph{{\bf Weight learning.}} {In applications, we usually consider LRNNs of the form $\mathcal{N} \cup \mathcal{E}$, where $\mathcal{N}$ is a weighted set of first-order rules and $\mathcal{E}$ is a weighted set of ground facts. In particular, each $\mathcal{E}$ represents an example, while $\mathcal{N}$ acts as a template for constructing feed-forward neural networks, with $\overline{\mathcal{N}\cup \mathcal{E}}$ being the network corresponding to example $\mathcal{E}$. While the weights of $\mathcal{E}$ are given, the weights of $\mathcal{N}$ typically need to be learned from training data, as follows.}

We are given a list of examples $\mathcal{E} = (\mathcal{E}^1, \dots, \mathcal{E}^m)$ where each $\mathcal{E}^j$ is a LRNN (typically containing only weighted ground facts), and a list of training queries $\mathcal{Q} = ( \{ (q_1^1, t_1^1),\allowbreak \dots,\allowbreak (q_{k_1}^1,t_{k_1}^1) \},\allowbreak \dots,\allowbreak \{ (q_1^m, t_1^m),\allowbreak \dots,\allowbreak (q_{k_m}^m, t_{k_m}^m) \} )$ where each $q_i^j$ is a ground atom, which we call a {\em training query atom}, and $t_i^j$ is its {\em target value}. 
For a query atom $q_{i}^j$, let $y_i^j$ denote the output of the atom neuron $A_{q_i^j}$ in the ground neural network of $\overline{\mathcal{N}\cup \mathcal{E}^j}$. The goal of the learning process is to find the weights $w_h$ of the rules (and possibly facts) in $\mathcal{N}$ for which the loss $J$ on the training query atoms $J(\mathcal{Q}) = \sum_{j = 1}^m \sum_{i = 1}^{k_j} \textit{loss}(y_i^j,t_i^j)$ is minimized. This loss function is then optimized using standard stochastic gradient descent algorithm \cite{sgd}. For details about weight learning of LRNNs, see~\cite{Sourek2015}.

\section{Structure Learning}\label{sec:struct}

In this section we describe a structure learning algorithm for LRNNs. The algorithm receives a list of training examples and a list of training queries, and it produces a LRNN. For simplicity, we will assume {that constants are only used as identifiers of objects. In particular, we will assume that attribute values are represented using unary literals, e.g.\ we would use} $\textit{red}(o)$ instead of $\textit{color}(o,\textit{red})$. Besides that we do not put any restrictions on the structure of the training examples.

\subsection{Structure of the Learned LRNNs}\label{sec:structure:structure}

The structure learning algorithm will create LRNNs having a generic ``stacked'' structure which we now describe. First, there are rules that define $d$ new predicates, representing {\em soft clusters} \cite{Sourek2016} of unary predicates from the dataset. These can be thought of as the first layer of the LRNN, where the weighted facts from the dataset comprise the zeroth layer.
For instance, if the unary predicates in the dataset are $A,B,\dots,Z$ then the LRNN will contain the following rules:
\begin{align*}
w_{a_1} &: \alpha^1_1(X) \leftarrow {A}(X)&
w_{b_1} &: \alpha^1_1(X) \leftarrow {B}(X)&
& ... &
w_{z_1} &: \alpha^1_1(X) \leftarrow {Z}(X)\\
w_{a_2} &: \alpha^1_2(X) \leftarrow {A}(X)&
w_{b_2} &: \alpha^1_2(X) \leftarrow {B}(X)&
& ... & 
w_{z_2} &: \alpha^1_2(X) \leftarrow {Z}(X)\\
 & ... & & ... & & ... & & ... &\\
w_{a_d} &: \alpha^1_d(X) \leftarrow {A}(X)&
w_{b_d} &: \alpha^1_d(X) \leftarrow {B}(X)&
& ... &
w_{z_d} &: \alpha^1_d(X) \leftarrow {Z}(X)
\end{align*}
\noindent Here each $\alpha^i_j$ is a latent predicate representing a soft cluster, the index $i$ denotes the layer in which it appears (in this case, the first layer) and $j$ indexes the individual soft clusters in that level.


In general, the second layer will consist of two types of rules. First, there may be rules introducing new latent predicates. In contrast to the unary predicates that were introduced in the first layer, here the latent predicates could be also of higher arity , although in practice an upper bound will be imposed for efficiency reasons. In the body of these rules, we may find predicates from the dataset itself, or latent predicates that were introduced in the first layer. The new latent predicates introduced in these rules may then be used in the bodies of rules in subsequent layers. Second, there may also be rules that have a predicate from the dataset in their head. These will typically be rules that were learned to predict the target predicates that we want to learn.


\begin{example}
For instance, in datasets of molecules, unary predicates can be used to represent types of atoms, such as {\em carbon} or {\em hydrogen}. 
An example of a possible second layer rule is:
$$
w_{p_1} : p_1(X,Y) \leftarrow \textit{bond}(X,Y) \wedge \alpha^1_1(X) \wedge \alpha^1_2(Y)
$$
Here $p_1$ is assumed to be one of the predicates from the dataset. 
Second layer rules that introduce a new latent predicate could look as follows.
\begin{align*}
w^2_{1,1}& : \quad \alpha^2_1(V1,V2) &\leftarrow \quad & \textit{bond}(V1,V2) \wedge \alpha^1_1(V1) \wedge \alpha^1_1(V2)&\\
w^2_{1,2}& : \quad \alpha^2_1(V1,{V3}) &\leftarrow \quad & \textit{bond}(V1,V2) \wedge \textit{bond}(V2,V3) \wedge \alpha^1_1(V1) \wedge \alpha^1_1(V3)&
\end{align*}
The actual intuitive meaning of the predicate $\alpha^2_{1}$ will depend on the weights $w^2_{1,1}$, $w^2_{1,2}$. For instance, if both are large enough, the (atom neurons corresponding to the) predicate will have high output whenever its arguments correspond to two atoms which are either one or two steps apart from each other in the molecule, and which have sufficiently high membership in the soft cluster $\alpha^1_1$.
\end{example}

Any higher layers have a similar structure to the second layer, where the $n^{\textit{th}}$ layer contains rules whose bodies only contain predicates from layers 0 to $n-1$, and whose heads either contain a target predicate or introduce a new latent predicate.


\subsection{Structure Learning Algorithm}

\begin{algorithm}[t]
\caption{General schema of structure learning}
	\begin{algorithmic}[1]
	    \State $\mathcal{E} \gets$ learning examples
	    \State $d \gets$ latent concepts' dimension
	    \State $\mathcal{W}, \mathcal{V}, \mathcal{R} \gets \varnothing$
	    \State {$\mathcal{R} \gets createLayer1Rules(\mathcal{E},d)$}    
	    \State $\mathcal{W} \gets initWeights(R)$           
		\State $(\mathcal{F},\mathcal{V}) \gets weightedFacts(\mathcal{E},R,W)$
				
		\While{$\neg StoppingCriterion$}
		    \State $bestRule \gets ruleLearning(\mathcal{F},\mathcal{V},\mathcal{R})$   
		    \State $bestRules \gets predicateInvention(bestRule)$            
		    \State $\mathcal{R} \gets \mathcal{R} \cup bestRules$
		    \State $\mathcal{W} \gets trainWeights(\mathcal{R},\mathcal{E},\mathcal{W})$

			\State $(\mathcal{F},\mathcal{V}) \gets weightedFacts(\mathcal{E},\mathcal{R},\mathcal{W})$   

		\EndWhile
		\State \Return $(\mathcal{R},\mathcal{W})$
	\end{algorithmic}
	\label{alg:structLearn}
\end{algorithm}

The structure learning algorithm (Algorithm \ref{alg:structLearn}) iteratively constructs LRNNs that have the structure described in the previous section. It alternates weight learning steps with rule learning steps\footnote{
Variants of this strategy are employed by many structure learning algorithms in the context of statistical relational learning, e.g. \cite{sayu,kok2005learning,dinh2011generative}.}. In the weight learning steps, the algorithm uses stochastic gradient descent to minimise the squared loss of the LRNN by optimising the weights of the rules, as described in Section \ref{sec:preliminaries}. In the rule learning steps, the algorithm fixes the weights of all rules which define latent predicates and it searches for some {\em good} rule $R$. This rule $R$ should be such that the squared loss of the LRNN decreases after we add $R$ to it and and after we retrain the weights of all rules with non-latent head predicates. Next we describe this algorithm in detail.

The first step of the structure learning algorithm (lines 4--5) is the construction of the first level of the LRNN, which defines the unary predicates representing soft clusters of object properties, as described in Section \ref{sec:structure:structure}.

After the first step, the algorithm repeats the following procedure for a given number of iterations or until no suitable rules can be found anymore. It fixes the weights of all rules defining latent predicates (line 6). Then it runs a beam search algorithm searching through the space of possible rules\footnote{The space of rules is defined by two user-specified constraints: maximum rule length and maximum number of variables in a rule.} (line 8). The scoring function which is used by the beam search algorithm is computed as follows. Given a rule $R$, the algorithm creates a copy of the current LRNN to which the given candidate rule $R$ is added. It then optimises the log-loss of this new LRNN (which corresponds to maximum-likelihood estimation for logistic regression), training just the non-fixed weights, i.e.\ the weights of the rules with non-latent predicates in their heads. The score of the rule $R$ is then defined to be the log-loss after training the non-fixed weights. The reason why we do not retrain all weights of the LRNN when checking score of a rule $R$ are efficiency considerations because training the weights of the whole LRNN corresponds to training a deep neural network.
After the beam search algorithm finishes, the rule $R^*$ that it returned is added to the original LRNN. 

Note that $R^*$ contains one of the target predicates in its head. However, in addition to adding $R^*$, we also add a set of related rules that have latent predicates in their head (line 9), as follows. Here, we will assume for simplicity that all latent predicates have the same arity $k$, but the same method can still be used when the latent predicates are allowed to have different arities. Let $i$ be the highest index such that $R^*$ contains a latent predicate of the form $\alpha_j^i$ (i.e.\ a latent predicate from layer $i$) in its body, where we assume $i=1$ if $R^*$ does not contain any latent predicates 
. Then for each latent predicate $\alpha^{i+1}_j$ from the $(i+1)$-th layer, the algorithm adds to the LRNN all rules which have $\alpha^{i+1}_{j}(V_1,\dots,V_k)$ in the head and which can be obtained by unifying $V_1,\dots,V_k$ with the variables in $R^*$. This process is illustrated in the following example.
\begin{example}
Revisiting the example of molecular datasets, let $R^* = p(A,B) \leftarrow \textit{bond}(A,B) \wedge \alpha^1_2(A) \wedge \alpha^2_5(B)$ and let $k = 1$. Then the algorithm will add the following latent-predicate rules:
\begin{align*}
w^3_{1,1}& : \quad \alpha^3_1(V_1) &\leftarrow \quad & \textit{bond}(V_1,B) \wedge \alpha^1_2(V_1) \wedge \alpha^2_5(B)&\\
w^3_{1,2}& : \quad \alpha^3_1(V_1) &\leftarrow \quad & \textit{bond}(A,V_1) \wedge \alpha^1_2(A) \wedge \alpha^2_5(V_1)&\\
w^3_{2,1}& : \quad \alpha^3_2(V_1) &\leftarrow \quad & \textit{bond}(V_1,B) \wedge \alpha^1_2(V_1) \wedge \alpha^2_5(B)&\\
w^3_{2,2}& : \quad \alpha^3_2(V_1) &\leftarrow \quad & \textit{bond}(A,V_1) \wedge \alpha^1_2(A) \wedge \alpha^2_5(V_1)&\\
\dots &  \quad \quad  \dots & & \dots &\\
w^3_{d,1}& : \quad \alpha^3_{d}(V_1) &\leftarrow \quad & \textit{bond}(V_1,B) \wedge \alpha^1_2(V_1) \wedge \alpha^2_5(B)&\\
w^3_{d,2}& : \quad \alpha^3_{d}(V_1) &\leftarrow \quad & \textit{bond}(A,V_1) \wedge \alpha^1_2(A) \wedge \alpha^2_5(V_1)&
\end{align*}
Note that the algorithm has to add the new rules to the layer $3$ because $R^*$ already contained predicates from the layer $2$.
\end{example}

 After the LRNN has been extended by all these rules obtained from $R^*$, the weights of all the rules, including those corresponding to latent predicates, are retrained using stochastic gradient descent (line 11). Note that typically there will be some latent predicates which are not used in any rules; their weights are not considered during training. Subsequently, the algorithm again fixes the weights of the rules corresponding to the latent predicates, and repeats the same process to find an additional rule. This is repeated until a given stopping condition is met.
 

\section{Experiments}\label{sec:experiments}

In this section we describe the results of experiments performed with the structure learning algorithm on a real-life molecular dataset.
We performed experiments on 72 NCI datasets \cite{ncigi}, each of which contains several thousands of molecules, labeled by their ability to inhibit the growth of different types of tumors. 
We compare the performance of the proposed LRNN structure learning  method with the best previously published LRNNs, which contain large generic, yet manually constructed weighted rule sets~\cite{Sourek2015}. For further comparison we include the relational learners kFOIL \cite{kFOIL} and nFOIL \cite{nFOIL}, which respectively combine relational rule learning with support vector machines and with naive Bayes learning. 

\begin{figure}[t]
\resizebox{\textwidth}{5.0cm}{
\includegraphics[trim={1.9cm 5.2cm 2cm 6cm},clip,width=\textwidth]{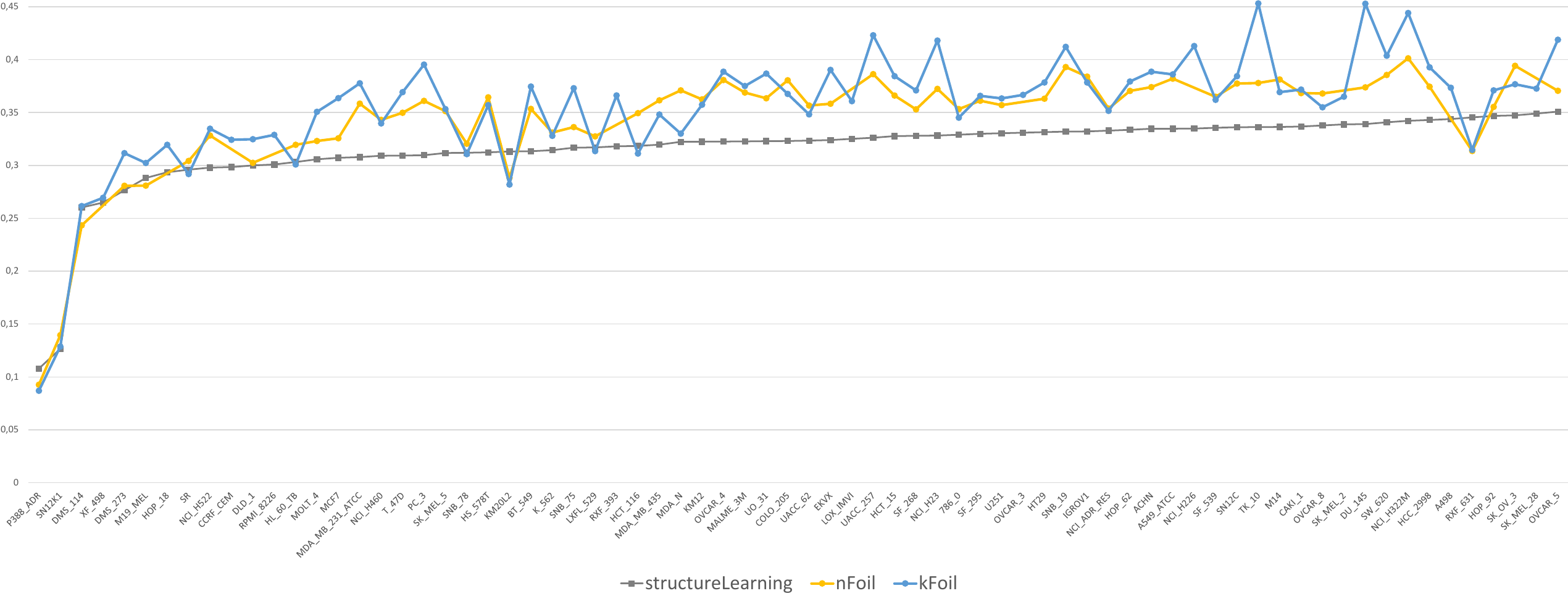}
}
\caption{Comparison of crossvalidated test errors of LRNNs produced by structure learning with nFoil and kFoil learners as baselines. 
}
\label{fig:results1}
\end{figure}

\begin{figure}[h]
\resizebox{\textwidth}{5.0cm}{
\includegraphics[trim={1.9cm 5.2cm 2cm 6cm},clip,width=\textwidth]{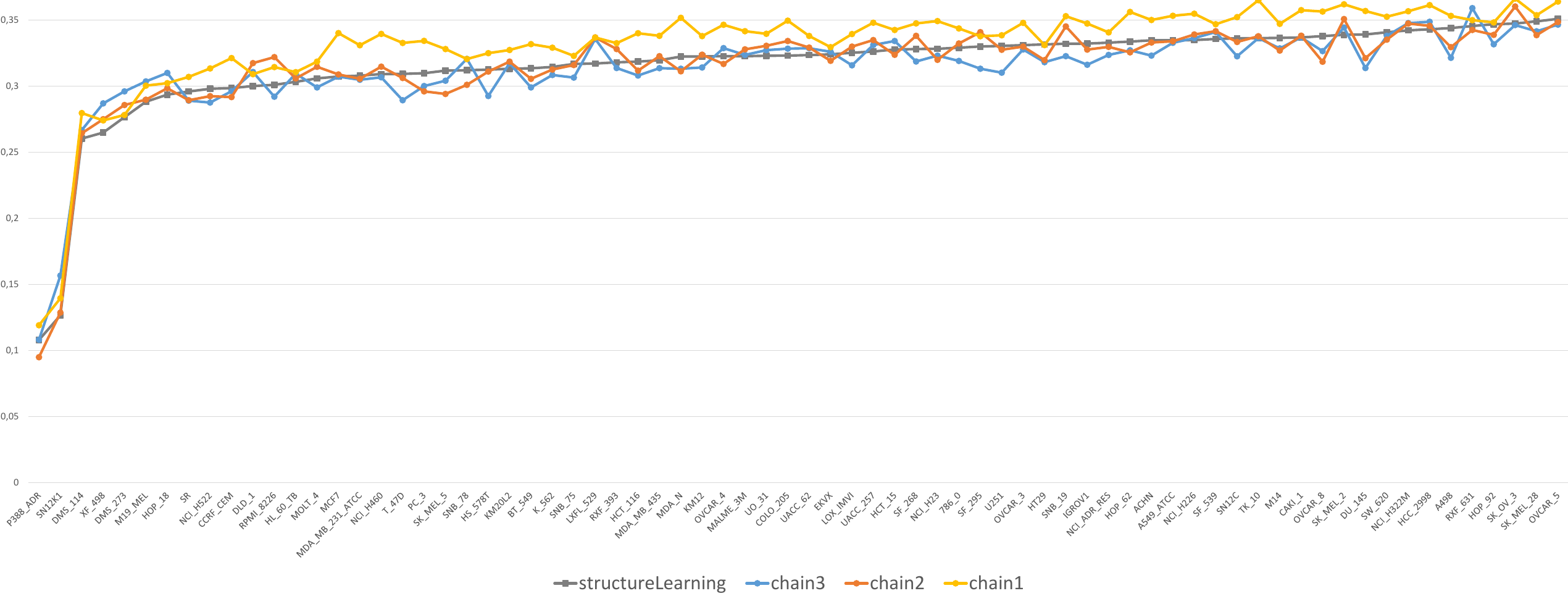}
}
\caption{Comparison of test errors of LRNNs produced automatically by structure learning with 3 handcrafted LRNNs with varying lengths of chain patterns from~\cite{Sourek2015}. 
}
\label{fig:results2}
\end{figure}

\begin{figure}[h]
\resizebox{\textwidth}{4cm}{
\includegraphics[trim={4cm 0cm 3.7cm 1cm},clip,width=\textwidth]{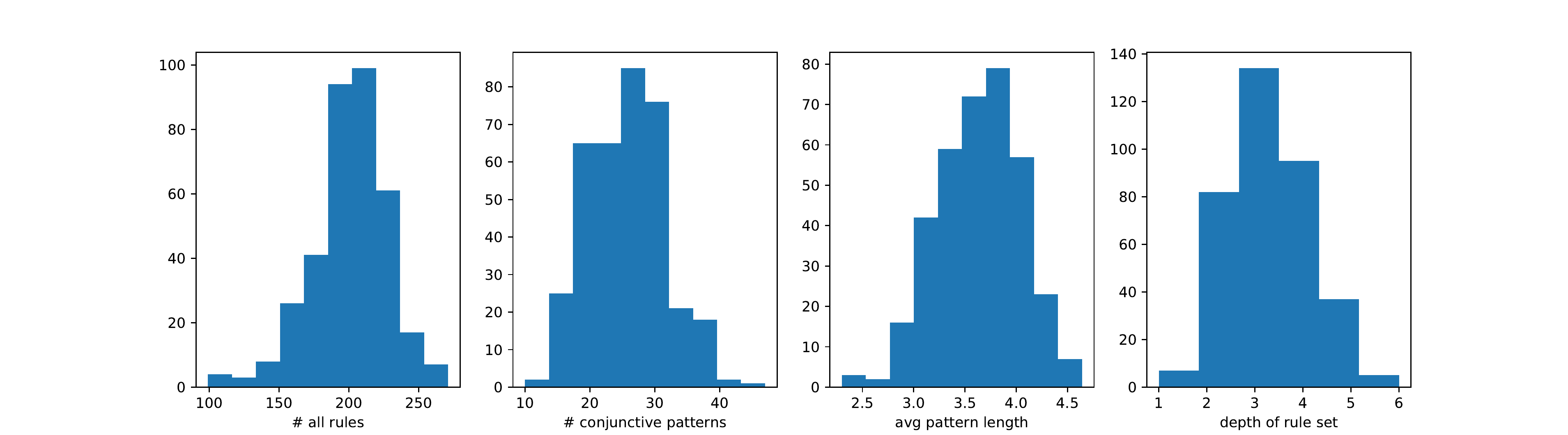}
}
\caption{Statistics of the learned LRNN rule sets from experiments with the 72 NCI datasets. We display (i)~the number of rules (including zeroth layer soft clusters), (ii)~the number of conjunctive rules (patterns) learned, (iii)~the average length of these rules (patterns),
and (iv)~the overall number of layers (depth of template).
}
\label{fig:templateStats}
\end{figure}

The results are shown in Figure~\ref{fig:results1} and Figure~\ref{fig:results2}. The automatically learned LRNNs outperform both kFOIL and nFOIL in terms of predictive accuracy (measured using cross-validation). The learned LRNNs are also competitive with the manually constructed LRNNs from \cite{Sourek2015a,Sourek2015}, although they do not outperform them. They are slightly worse than the largest of the manually constructed LRNNs, based on graph patterns with $3$ vertices, enumerating all possible combinations of soft cluster types of the three atoms and {soft cluster types} of the two bonds connecting them. Figure \ref{fig:templateStats} displays statistics of the learned LRNN rule sets. These statistics show that the structure learner turned out to produce quite complex LRNNs having multiple layers of invented latent predicates.
 
\begin{figure}
\resizebox{\textwidth}{5.0cm}{
\includegraphics[trim={1.5cm 0.9cm 0cm 0.2cm},clip,width=\textwidth]{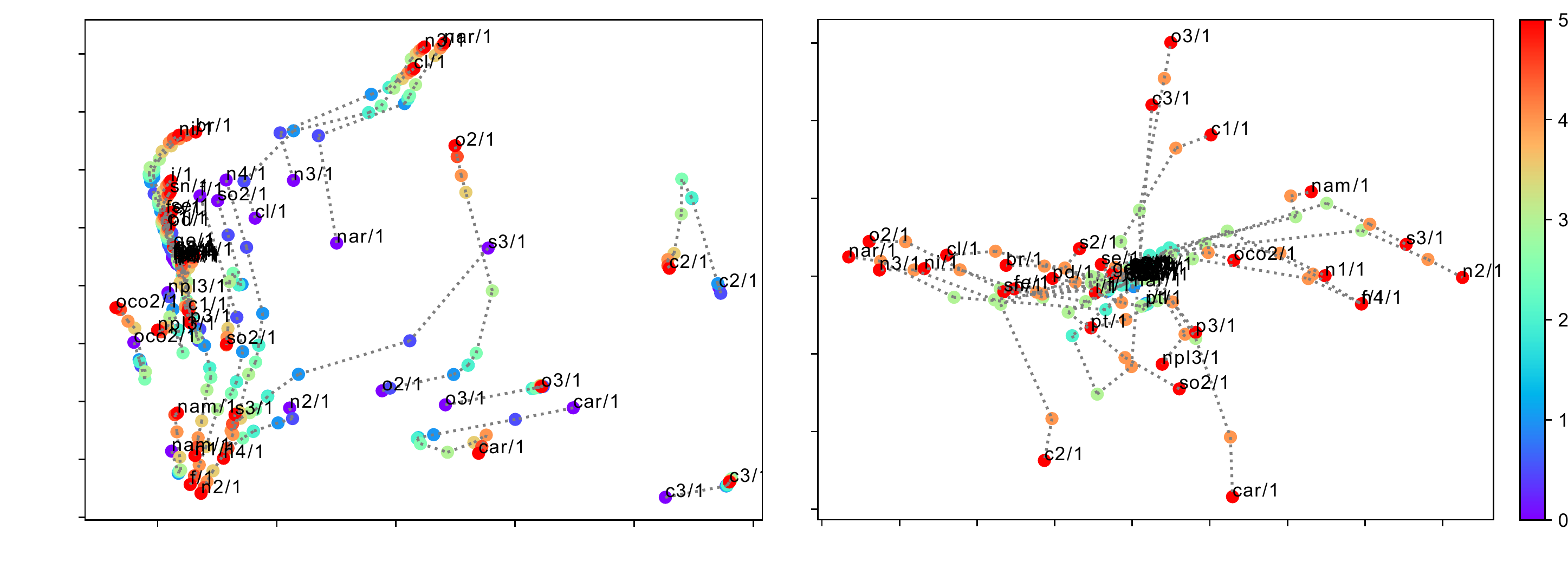}
}
\caption{PCA projection of evolution of atom embeddings during first 6 iterations (denoted by colors) of structure learning of a LRNN, with initialization based on unsupervised pre-training (left) and with completely random initialization (right).}
\label{fig:embedding}
\end{figure}

The weights of the rules defining the latent predicates in the first layer of the LRNN can be interpreted as coordinates of a vector-space embedding of the properties (atom types in our case). In Figure \ref{fig:embedding}, we plot the evolution of these embeddings as new rules are being added by the structure learning algorithm. The left panel of Figure \ref{fig:embedding} displays the evolution of the embeddings of atom types after these have been pre-trained using an unsupervised method which was originally used for statistical predicate invention in \cite{Sourek2016}. The right panel of the same figure displays the evolution of the embeddings when starting 
from random initialization without any unsupervised pre-training. What can be seen from these figures is how, as the model becomes more complex, the atom types start to make more visible clusters. 
Interestingly and perhaps somewhat against intuition, the use of the unsupervised pre-training seemed to consistently decrease predictive performance (we omit details due to limited space).

\section{Related Work}\label{sec:related}

LRNNs are related to many older works on using neural networks for relational learning such as~\cite{blockeel2004using} and more recent approaches such as \cite{rocktaschel2016learning,cohen2016tensorlog}.
The structure learning strategy that we employ in the methods presented in this paper is in many respects similar to structure learning methods from statistical relational learning such as \cite{sayu,kok2005learning,dinh2011generative}. However, what clearly distinguishes it from all these previous SRL approaches is its ability to automatically induce hierarchies of latent concepts. In this respect, it is also related to meta-interpretive learning \cite{DBLP:journals/ml/MuggletonLT15}. However, meta-interpretive learning is only applicable to the learning of crisp logic programs. The structure learning approach is also related to works on refining architectures of neural networks~\cite{fahlman1989cascade,opitz1993heuristically}. However, from these it differs in its ability to handle relational data.


\section{Conclusions and Future Work}\label{sec:conclusion}

In this paper we have introduced a method for learning the structure of LRNNs, capable of learning deep weighted rule sets with invented latent predicates. The predictive accuracies obtained by the learned LRNNs were competitive with results that we obtained in our previous work using manually constructed LRNNs. The method presented in this paper therefore has the potential to make LRNNs useful in domains where it would otherwise be difficult to come up with a rule set manually. It also makes the adoption of LRNNs by non-expert users more straightforward, as the proposed method can learn competitive LRNNs without requiring any user input (besides the dataset).

\vspace{0.2cm}
{\small \noindent {\bf Acknowledgements} G\v{S}, MS and F\v{Z} acknowledge support by project no.\ 17-26999S granted by the Czech Science Foundation. OK is supported by a grant from the Leverhulme Trust (RPG-2014-164). SS is supported by ERC Starting Grant 637277.
Computational resources were provided by the CESNET LM2015042 and the CERIT Scientific Cloud LM2015085, provided under the programme ``Projects of Large Research, Development, and Innovations Infrastructures".}

\bibliographystyle{spmpsci}      
\bibliography{main}   

\end{document}